\pdfoutput=1
\documentclass{ntu_glasso}
\usepackage{tikz}
\usepackage{mathtools}
\usepackage{graphicx}
\usepackage[square,sort,comma,numbers]{natbib}
\usepackage{enumerate}
\usepackage{subfigure}
\usepackage{caption}
\usepackage{dsfont}

\usepackage{xcolor,colortbl}
\usepackage{hyperref}
\definecolor{Gray}{gray}{0.85}
\newcommand{\mc}{\mathcal}
\newcommand{\mb}{\mathbb}
\makeatletter
\def\@copyrightspace{\relax}
\makeatother
\begin{document}

\title{Segmentation Analysis in Human Centric Cyber-Physical Systems using Graphical Lasso}

\numberofauthors{6}

\author{
\alignauthor
Hari Prasanna Das
\thanks{Both authors contributed equally.}
\thanks{Corresponding Author: \textit{Hari Prasanna Das}}\\
      \affaddr{UC Berkeley}\\
      \affaddr{Berkeley, California 94720}\\
      \affaddr{hpdas@berkeley.edu}
\alignauthor
Ioannis C. Konstantakopoulos
\footnotemark[1]\\
      \affaddr{UC Berkeley}\\
      \affaddr{Berkeley, California 94720}\\
      \affaddr{ioanniskon@berkeley.edu}
\alignauthor
Aummul Baneen Manasawala\\
      \affaddr{UC Berkeley}\\
      \affaddr{Berkeley, California 94720}\\
      \affaddr{abf@berkeley.edu}
\and  
\alignauthor
Tanya Veeravalli\\
      \affaddr{UC Berkeley}\\
      \affaddr{Berkeley, California 94720}\\
      \affaddr{tveeravalli19@berkeley.edu}
\alignauthor
Huihan Liu\\
      \affaddr{UC Berkeley}\\
      \affaddr{Berkeley, California 94720}\\
      \affaddr{liuhh@berkeley.edu}
\alignauthor
Costas J. Spanos\\
      \affaddr{UC Berkeley}\\
      \affaddr{Berkeley, California 94720}\\
      \affaddr{spanos@berkeley.edu}
}
\maketitle

\begin{abstract}
A generalized gamification framework is introduced as a form of smart infrastructure with potential to improve sustainability and energy efficiency by leveraging humans-in-the-loop strategy. The  proposed framework enables a  \textit{Human-Centric Cyber-Physical System} using an interface to allow building managers to interact with occupants. The interface is designed for occupant engagement-integration supporting learning of their preferences over resources in addition to understanding how preferences change as a function of external stimuli such as physical control, time or incentives. Towards intelligent and autonomous incentive design, a noble statistical learning algorithm performing occupants energy usage behavior segmentation is proposed. We apply the proposed algorithm, Graphical Lasso, on energy resource usage data by the occupants to obtain feature correlations---dependencies. Segmentation analysis results in characteristic clusters demonstrating different energy usage behaviors. The features-factors characterizing human decision-making are made explainable.
\end{abstract}

\section{Introduction}
Energy consumption of buildings, both residential and commercial, account for approximately 40\% of all energy usage in the U.S.~\cite{mcquade2009}. In efforts to improve energy efficiency in buildings, researchers and industry leaders have attempted to implement control and automation approaches alongside techniques like incentive design and price adjustment to more effectively regulate the energy usage. A building manager, acting as the connection between energy utilities and the end users, can be motivated to encourage energy-efficient behaviors in many ways. 
The most obvious requirement is to maintain an energy efficient building with operational excellence. We consider the design of an occupant-interactive framework with incentive design for improving energy efficiency as the base of this research work.

We have dual objectives. Our first objective is to improve building energy efficiency by introducing a gamification system that engages users in the process of energy management and integrates seamlessly through the use of cyber-physical technology by leveraging humans-in-the-loop strategy. \textit{Human-Centric Cyber-Physical System} has a great potential not only as a design component of noble smart building architectures~\cite{jia2018design,biscuit,Liu_thermal_comfort}, but also as a part of broader \textit{societal-scale cyber-physical systems}~\cite{ratliff2015incentivizing}. There exists a considerable amount of previous work on the success of control and automation in the improvement of building energy efficiency~\cite{Aswani:2012kx,boman:1998aa}. Some notable techniques with encouraging results implement concepts such as incentive design and adaptive pricing~\cite{Dahleh2010smartCom}. 

Specifically, control theory has been a critical source for several approaches that employ ideas like model predictive and distributed control and have demonstrated encouraging results in applications like Heating, Ventilation and Air Conditioning (HVAC). Unfortunately, the control approaches which are applied to human-centric environments lack the ability to consider the individual preferences of occupants. This trend is also apparent in machine learning approaches to HVAC system control. While these approaches are capable of generating optimal control designs, they fail to adjust to occupant preferences and the consequences of their presence in the system. The heterogeneity of user preferences in regard to building utilities is considerable in variety and necessitates a system that can adequately account for differences from one occupant to another.

Clearly, the presence of occupants considerably complicates the determination of an efficient building management system. With this in mind, focus has shifted towards modeling occupant behavior within the system in an effort to incorporate their preferences. To accomplish this task, the building and its occupants are represented as a multi-agent system targeting occupant comfort~\cite{boman:1998aa}. First, occupants and managers are allowed to express their building preferences. Then the preferences are used to generate an initial control policy. An iteration on this control policy is created by using a rule engine that attempts to find compromises between the preferences. Some drawbacks of this control design are immediately apparent. There is no form of communication to the manager about occupant preferences. In addition, there exists no incentive for submission of true user preferences, and no system is in place for feedback from occupants. Other related topics in the same vein focus on grid integration~\cite{samad:2016aa} while still others consider approaches for policy recommendations and dynamic pricing systems \cite{Dahleh2010smartCom}.

Our second objective is to make the incentive design process more intelligent and autonomous by learning the factors leading to human decision-making, and using the knowledge to devise a noble agent segmentation method. Customer segmentation has been a well-studied area in energy systems, as energy utility companies frequently use segmentation techniques for optimal planning of demand response, load shedding, and microgrid applications among others~\cite{energy_utility}. The optimal incentive design for a game in \textit{Human-Centric Cyber-Physical Systems} with high dimensional data would require powerful yet computationally efficient statistical methods. A possible candidate, Graphical Lasso algorithm, has been widely applied on different scientific studies due to its sparsity property ($\ell_1$ penalty term) and efficiency \cite{glasso_biology,cvpr_scene,HallacPBL17}. To demonstrate the efficacy of the above factors, i.e. the potential of Graphical Lasso and the importance of customer segmentation in the energy industry, we enable new avenues by combining both concepts and applying on a gamification---social game data set to classify the energy efficiency behaviors among building occupants. We explore the causal relationship between different features of the agents using a versatile tool, Grangers Causality, which leads to a deep understanding of decision-making patterns and helps in integrating explainable game theory models with adaptive control or online incentive design.

With the advent of explainable Artificial Intelligence, there has been a massive move towards making statistical models explainable. Our proposed method is explainable, rather than being just a black box model. 

To summarize, our contributions are threefold:

\begin{itemize}
    \item Implementation of a large-scale gamification application with the goal of improving energy  efficiency in buildings. 
    \item Novel segmentation analysis using an explainable statistical model at the core towards learning agents (building occupants) contributed features---factors characterizing their decision-making in competitive environments.
    \item Characterization of causal relationship among several contributed features---factors explaining decision-making patterns in agents' actions.
\end{itemize}
\section{Related Work}

In order to place the work pertaining to cyber-physical systems in the context of the state of the art, we briefly overview existing approaches. As alluded to previously, the key to our approach is the implementation of a social game--a gamification framework among users in a non-cooperative setting that engages them in the process of energy management and integrates seamlessly through the use of human-centric cyber-physical technology. Similar methods that employ \textit{social games} have been applied to transportation systems with the goal of improving flow~\cite{merugu:2009aa,pluntke2013insinc}. Another example application can be found in the medical industry in the context of privacy concerns versus expending calories~\cite{bestick:2013aa}. Entrepreneurial ventures have also sought to implement solutions of their own to the problem of building energy efficiency. \textit{Comfy}\footnote{ https://comfyapp.com} and \textit{Cool Choices}\footnote{ https://coolchoices.com/how-it-works/improve} are two examples of start-ups that have developed interesting approaches to controlling building utilities.
 
 The idea behind the social game context is to create a friendly competition between occupants. In turn, this competition will help motivate them to individually consider their own energy usage and, hopefully, seek to improve it. The same technique of gamification has been used as a way to educate the public about energy usage~\cite{knol:2011aa,salvador:2012aa,orland:2014aa}. It has also been cleverly implemented in a system that presents feedback about overall energy consumption to occupants~\cite{simon:2012aa}. One case of a gamification methodology was used to engage individuals in demand response (DR) schemes~\cite{li:2014aa}. Each user is represented as a utility maximizer within the model of a Nash equilibrium where occupants gain incentives for reduction in energy consumption during DR events. In contrast to approaches that target user devices with known usage patterns~\cite{li:2014aa}, our approach focuses on personal room utilities such as lighting, without any initial usage information, simulating scenarios of complete ignorance to occupant behaviors. 
 
 Leveraging past users' observations, we can learn the utility functions of individual occupants by the way of several noble algorithms~\cite{konstantakopoulos2018deep}. In previous work, we have explored utility learning and incentive design as a coupled problem both in theory~\cite{konstantakopoulos2018robust,konstantakopoulos2016inverse} and in practice~\cite{konstantakopoulos2015social} under a Nash equilibrium approach. Through this approach, we can generate excellent prediction of expected occupant actions. Our unique social game methodology simultaneously learns occupant preferences while also opening avenues for feedback. This feedback is translated through individual surveys that provide opportunities to influence occupant behavior with adaptive incentive. With this technique, we are capable of accommodating occupant behavior in the automation of building energy usage by learning occupant preferences and applying a variety of noble algorithms. Furthermore, the learned preferences can be adjusted through incentive mechanisms to enact improved energy usage.
 
The broader purpose of this paper is to design a segmentation analysis for human-centric cyber-physical systems. Our novel segmentation analysis supports learning agent's contributed features---factors characterizing their decision-making in competitive environments such as the smart building energy social game described in section~\ref{social_game}. Most importantly, our method can be leveraged in the design of incentive mechanisms that realign agent's preferences with those of the planner. Advanced incentive---mechanism design algorithms~\cite{ratliff2018adaptive} are important in sustaining gamification applications and transforming them to a profitable application. Hence, our method serves to understand key features---factors which contribute to specific actions by users. 

Towards this we use high dimensional real-world data. More concretely, we use the graphical lasso algorithm as a powerful tool to understand the latent conditional independence between variables~\cite{mainbook}. This in turn provides insights into how different features interplay among each other. Historically, Graphical Lasso has been used in various fields of science, like how individual elements of the cell interact with each other~\cite{glasso_biology} and in the broad area of computer vision for scene labelling~\cite{cvpr_scene}. A modified version of the original algorithm, named time-varying graphical lasso, has been used on financial and automotive data~\cite{HallacPBL17}. However, the novelties of graphical lasso has not been well utilized in the area of cyber-physical systems. 

We derive inspiration for agent segmentation owing to the fact that customer segmentation has been successfully utilized in energy systems~\cite{energy_utility,automated_segmentation}. We use Granger causality to explain the causal relationship between the features in energy usage behavior of agents in social game. Granger causality is a statistical test used to determine causal relationship between two signals. If signal X granger-causes signal Y, then past values of X can be used to predict Y for future timesteps beyond what is available for Y. It has been widely used in the energy domain in applications such as deducing the causal relationship between economic growth and energy consumption~\cite{chiou2008economic}.

Bridging the gap between noble segmentation algorithms and their application to energy, we employ graphical lasso algorithm for customer segmentation on social game dataset and come up with an explainable model, helpful both in understanding inherent factors leading to energy efficiency in \textit{Human-Centric Cyber-Physical Systems} and incentive design for similar game theory applications.
\section{Gamification in Human Centric Cyber-Physical Systems}

\begin{figure}[!ht]
\centering
  \includegraphics[width=0.35\textwidth]{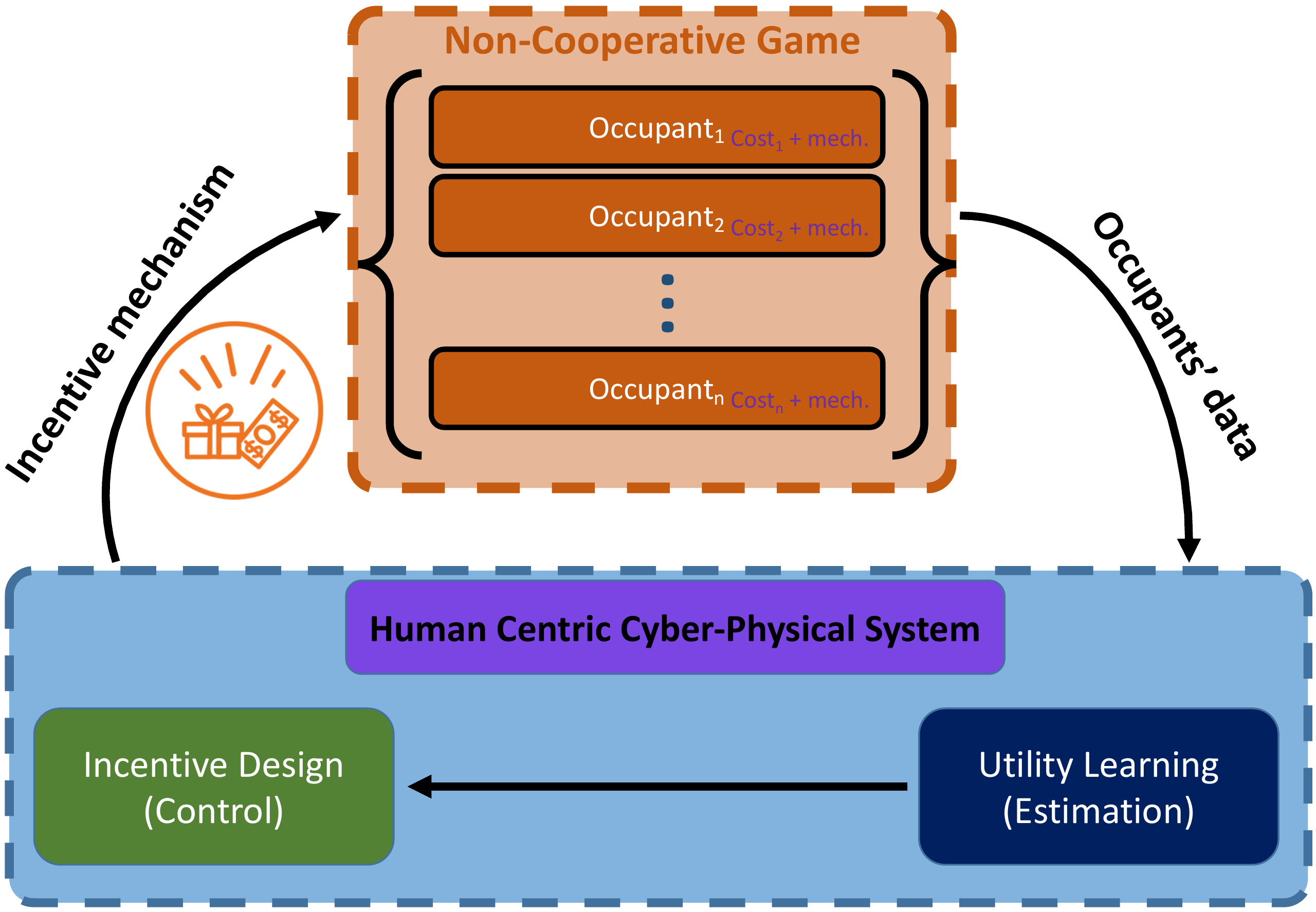}
  \caption{Gamification abstraction for \textit{Human Centric Cyber-Physical Systems} towards building energy efficiency: Iterative Utility Learning (Machine Learning \& Segmentation Analysis) and Incentive Design}
  \label{fig:gamification_abstraction}
\end{figure}

We envision smart-building systems in which humans take control, interact, and improve the environment they inhabit. Human decision-making plays a critical role in the management and operation of contemporary infrastructure, due in no small part to the advent of Internet of Things and cyber-physical sensing/actuation platforms in industry. Adoption of human-centric building services and amenities also leads to improvements in the operational efficiency of cyber-physical systems. 

However, human preference in regard to living conditions is usually unknown and heterogeneous in its manifestation as control inputs to a building. Furthermore, the occupants of a building typically lack the independent motivation necessary to contribute to and play a key role in the control of smart building infrastructure. Hence, it poses a potential for an efficient application of game theory. We introduce a generalized gamification abstraction (Figure ~\ref{fig:gamification_abstraction}) towards enabling strategic interactions among non-cooperative agents in \textit{Human-Centric Cyber-Physical Systems}. A Human-centric cyber-physical system can be defined as

\begin{center}
\textit{A system that combines the cyber-world (computer-based technologies) with several physical processes in which humans are directly integrated and are interacting with the environment through sensing/actuation platforms and control strategies.}
\end{center}

The proposed framework supports learning agents preferences over shared or scarce resources as well as understanding how preferences change as a function of external stimuli such as physical control, time or incentives. Our gamification framework is designed to support engagement and integration in a human-centric cyber-physical system. The broader purpose of this section is to present a general framework that leverages game-theoretic concepts to learn models of player's decision making in competitive environments such as the building energy social game.

\subsection{Game theoretic framework}\label{game_model}

In this section, we abstract the agents' decision making processes in a game-theoretic framework and specifically introduce discrete choice theory. Discrete choice theory~\cite{train2009discrete} is greatly celebrated in the literature as a means of data-driven analysis of human decision making. Under a discrete choice model, the possible outcome---decision of an agent can be predicted from a given choice set using a variety of available features. In section~\ref{graph_lasso}, we propose is quite broad method for characterizing features' contribution---structure to decision-making process.

\subsubsection{Agent decision-making model}

Consider an agent $i$ and the decision-making choice set which is mutually exclusive and exhaustive. The decision-making choice set is indexed by the set $\mc{I}=\{\mc{J}^1,\ldots, \mc{J}^S \}$. Decision maker $i$ chooses between $S$ alternative choices and would earn a \textbf{representative utility} $f_i$ for $i \in \mc{I}$. Each decision among decision-making choice set leads agents to get the highest possible utility, $f_i > f_j$ for all $i,j \in \mc{I}$. In our setting, an agent has a utility which depends on a number of features $x_z$ for $z=1, \ldots, N$. However, there are several unobserved components---features of the representative utility which should be treated as random variables. Hence, we define a \textbf{random utility} decision-making model for each agent given by 

\begin{equation}
  \hat{f}_i(x)=g_i(\beta_{i},x) + \epsilon_{i}
  \label{eq:discrete_utility}
\end{equation}

where $\epsilon_{i}$ is the unobserved random component of the agent's utility, $g_i(\beta_{i},x)$ is a nonlinear generalization of agent $i$'s utility function where

\begin{equation}
x=(x_1, \ldots, x_{i-1}, x_{i+1}, \ldots, x_N)\in \mb{R}^{N} 
\label{eq:feature}
\end{equation} 

is the
collective $n$ features explaining an agent's decision process. The choice of nonlinear mapping $g_i$ and $x$ abstracts the agent's decision. This is an adaptation of their classical representation~\cite{train2009discrete} using a linear mapping $g_i(\beta_{i},x) = \beta_{i}^{T}x$ in which $\epsilon_{i}$ is an independently and identically distributed random value modeled using a Gumbel distribution. In general, each agent is modeled as a \emph{utility maximizer} that seeks to select $i\in \mc{I}$ by optimizing \eqref{eq:discrete_utility}.


\subsubsection{Game formulation}

To model the outcome of the strategic interactions of agents we use a \emph{sequential non-cooperative discrete game} concept. We enable a more general framework of the discrete choice theory including a temporal dependence term, which supports modeling of stationary or non-stationary data---equilibrium points resulting from sequential equilibrium concept. Introducing a generalized decision-making model for each agent~\eqref{eq:discrete_utility}, in which random utility can be modeled either with linear or nonlinear mapping, a sequential non-cooperative discrete game is given as following.


Each agent $i$ has a set $\mc{F}_i = {f_{i}^1,\ldots, f_{i}^{N}}$ of $N$ \textbf{random utilities}. Each random utility $j$ has a convex decision-making choice set $\mc{I}_{j}=\{\mc{J}^{1}_{j},\ldots, \mc{J}^{S}_{j} \}$. Given a collection of $n$ features~\eqref{eq:feature} comprising the decision process, where the temporal parameter $T$ is also given, agent $i$ faces the following optimization problem for its \textbf{aggregated random utility}:

\begin{equation}
  \max\{\sum_{i=1}^{N}f^{T}_i(x)|\  f_i \in \mc{F}_i\}.
  \label{eq:opt-seq}
\end{equation}

In the sequential equilibrium concept agents in the game independently co-optimize their aggregated random utilities~\eqref{eq:opt-seq} given a collective of $n$ features~\eqref{eq:feature} at each time instance $T$ (temporal parameter). The above definition extends the definition of a discrete choice model~\cite{train2009discrete} to sequential games in which agents concurrently co-optimize several discrete (usually mutually exclusive) choices over temporal dependencies. In results section~\ref{results}, we present the dependencies of the decision-making process on temporal features based on the proposed graphical representation method~\ref{graph_lasso}.

\subsection{Human-centric CPS experiment}\label{social_game}


\begin{figure*}[!ht]
\center    \subfigure[\label{fig:game_utilization}]{\includegraphics[width=0.38\textwidth]{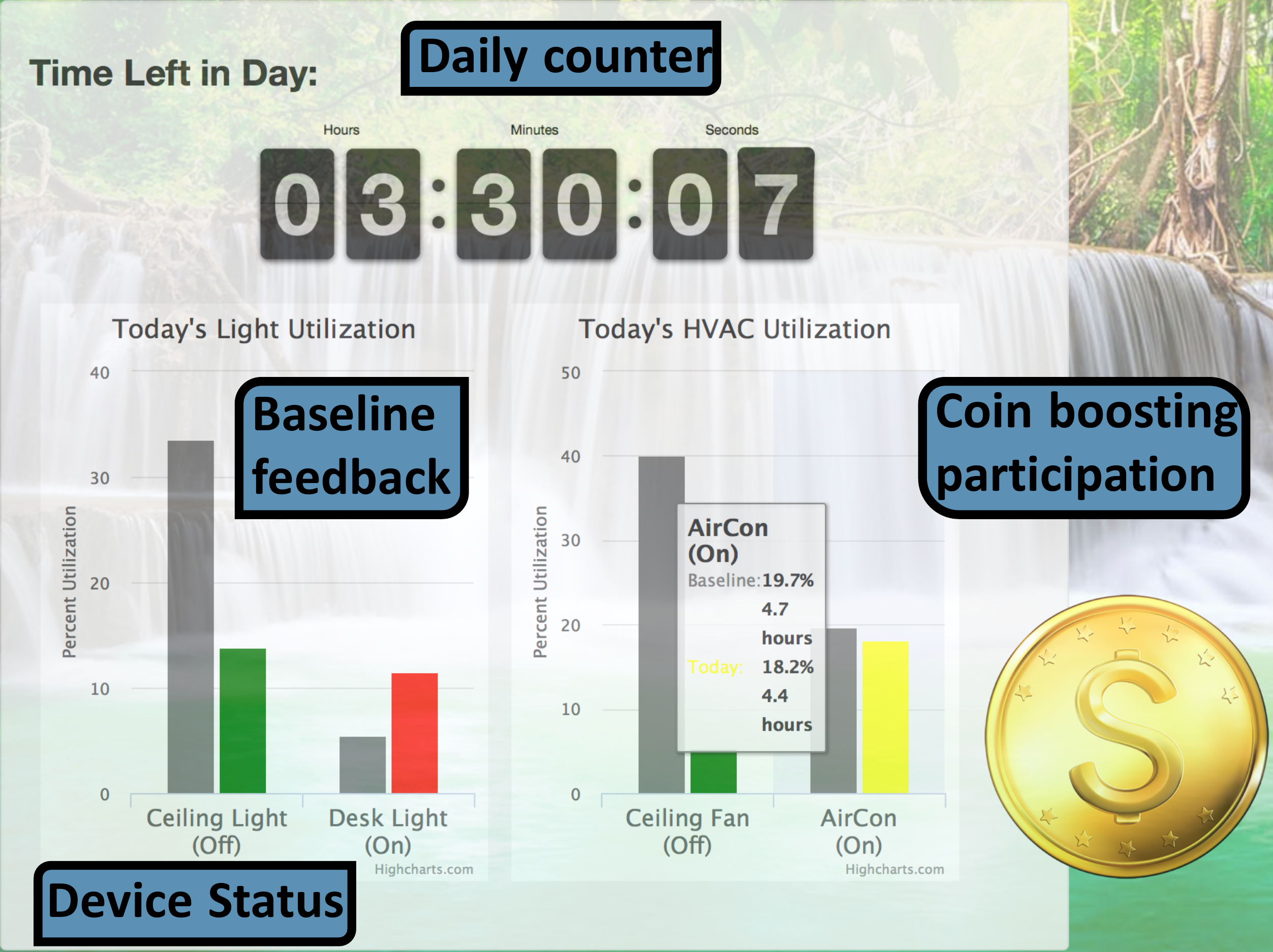}}
    \subfigure[\label{fig:IoT_design}]{\includegraphics[width=0.407\textwidth]{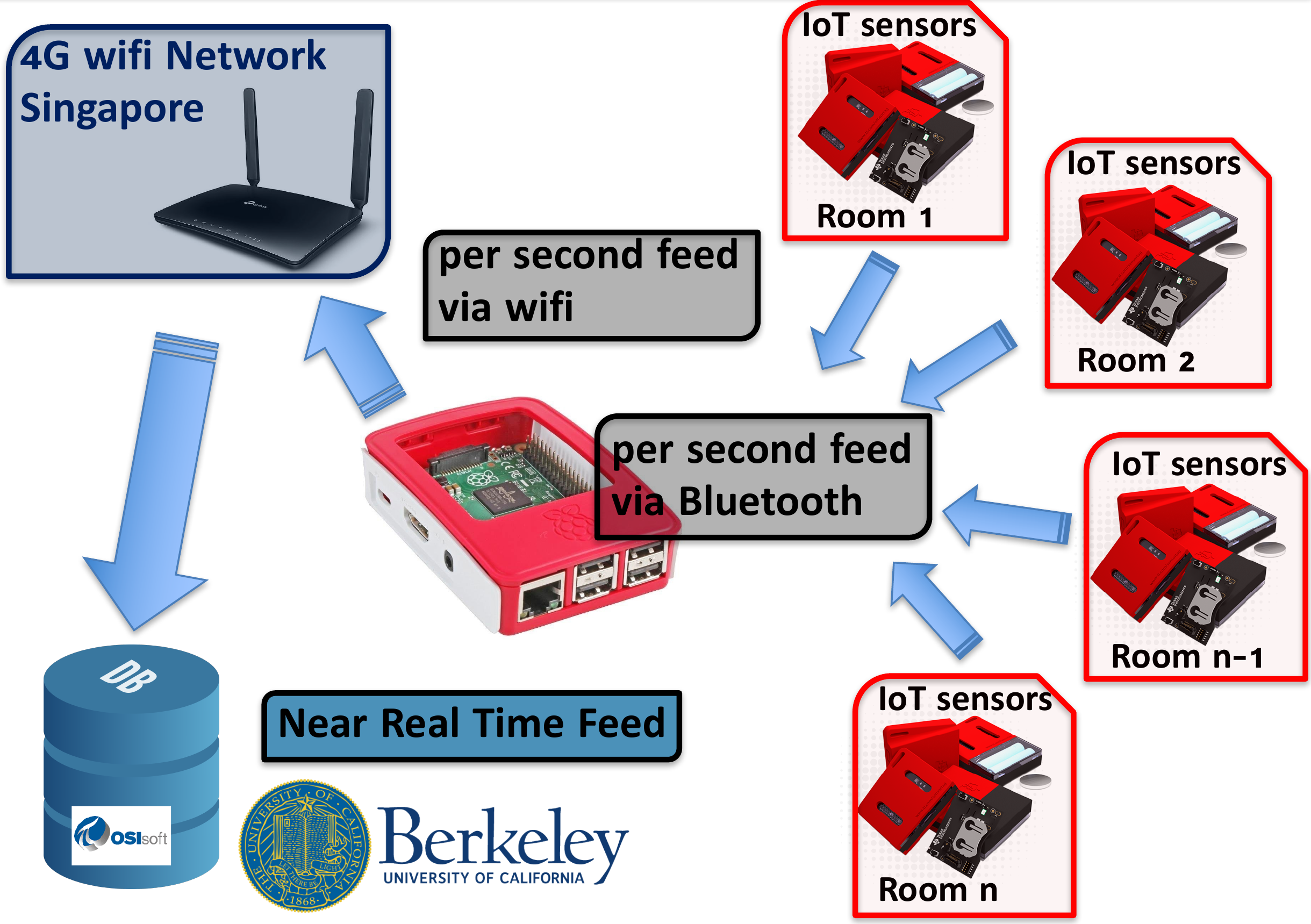}}
  \caption{Graphical user interface (GUI) and dataflow design for energy-based social game}
  \label{fig:game_utilization_design}
\end{figure*}


We briefly describe a social game experiment to encourage energy efficient resource consumption in a smart residential housing.\footnote{We refer interested readers to this paper~\cite{konstantakopoulos2018deep}, which includes the details of the experiment. Also, for demonstrations of our infrastructure and for downloading de-identified high dimensional data sets, please visit our web site: \href{url}{https://smartntu.eecs.berkeley.edu}}. We designed a social game which included residential housing single room apartments at Nanyang Technological University (NTU) campus. All single room dorm occupants were given access to a personal web-portal account---a convenient interface which displays their room's daily resource usage. Each of the dorm room was equipped with two Internet of Things (IoT) sensors\footnote{\textit{IoT sensor Tag}: \href{url}{http://www.ti.com}}---one close to the desk light and another near the ceiling fan. Deployed IoT sensors allowed dorm occupants to monitor their room's lighting system (desk and ceiling light usage) in real-time and HVAC (ceiling fan and A/C usage) with a refresh interval of up to one second. 

Monitoring past energy usage data for approximately one month before the introduction of the game for each semester, we calculated the baselines for each of the occupant's resources. Using this prior data, a weekday and weekend baseline was assigned. We accumulated data separately for weekdays and weekends so as to maintain fairness for occupants who have alternative schedules of occupancy (e.g. those who tend to stay at their dorm room over the weekends versus weekdays). Then, after introduction of the social game, dorm occupants were rewarded with points based on how energy efficient their daily usage is in comparison to their baseline usage.  We employed a lottery mechanism consisting of gifts awarded on a bi-weekly basis to incentivize occupants; occupants with more points are more likely to win the lottery. Earned points for each resource is given by

\begin{equation}
  \hat{p}^{d}_{i}(b_i, u_{i})= s_{i} \frac{b_i - u^{d}_{i}}{b_i}
  \label{eq:points_earned}
\end{equation}

where $\hat{p}^{d}_{i}$ is the points earned at day $d$ for room's resource $i$ which corresponds to ceiling light, desk light, ceiling fan, and A/C. Also, $b_i$ is the baseline calculated for each resource $i$, $u^{d}_{i}$ is the usage of the resource at day $d$, and $s_{i}$ is a points booster for inflating the points as a process of framing \cite{tversky1981framing}.

In Figure~\ref{fig:game_utilization_design}, we present how our graphical user interface was capable of reporting to occupants the real-time status (on/off) of their devices, their accumulated daily usage, time left for achieving daily baseline, and the percentage of allowed baseline being used by hovering above their utilization bars. In order to boost participation, we introduced a randomly appearing coin close to the utilization bars with the purpose of incentivizing occupants to log in to web-portal and view their usage. The residential housing single room apartments on the Nanyang Technological University campus were divided into four blocks, each of which had two levels. In this space, there were a total of seventy-two occupants who were eligible to participate in the social game. Participation in our social game platform was voluntary. We ran the experiment in both the Fall 2017 (September 12th - December 3rd) and Spring 2018 (February 19th - May 6th) semesters.

We enabled the design and implementation of a large-scale networked social game through the utilization of cutting-edge Internet of Things (IoT) sensors, advanced computational hardware devices (Raspberry Pi) and local weather monitoring station---systems (as an external parameter for our model). The actual data-feed was implemented through enabled 4G Wi-Fi Network in Singapore followed by posting of data(with minimal latency) to OSIsoft PI database \footnote{\textit{OSIsoft PI database}: https://www.osisoft.com} located at the University of California, Berkeley campus. The actual design and dataflow is depicted in Figure~\ref{fig:game_utilization_design}. 

Leveraging several indoor metrics like indoor illuminance, humidity, temperature, and vibrations (ceiling fan sensor), we derived simple thresholds indicating if a resource is in use or not. For instance, the standard deviation of acceleration derived from the ceiling fan mounted sensor is an easy way to determine whether ceiling fan is in the on state. Our calibrated detection thresholds are robust over daylight patterns, external humidity/temperature patterns, and noisy measurements naturally acquired from IoT sensors. While we were getting streamed data from various sensors in all dorm rooms, our back-end processes updated the status of the devices in near real-time in each occupant's account and updated points based on their usage and points calculation formula \eqref{eq:points_earned}. This functionality allowed occupants to receive feedback for their actions and view their points balance and their ranking among other capabilities.

\subsection{Social game data set}\label{social_game_data_set}

As a final step, we aggregated occupant's data in per-minute resolution. The per-minute features include time-stamp, each resource's status, accumulated resource usage (in minutes) per day, resource baseline, gathered points (both from game and surveys), occupant rank in the game over time and number of occupant's visits to the web portal. In addition to these features, we add several external weather metrics like humidity, temperature, and solar radiation among others. We pool more features utilizing a subset of the already derived features by leveraging domain knowledge and external information.\footnote{For downloading the described de-identified high dimensional data set, please visit our web site: \href{url}{https://smartntu.eecs.berkeley.edu}} Specifically, we consider \textbf{college schedule dummy feature indicators} including dummy variables for dates related to breaks, holidays, midterm and final exam periods at Nanyang Technological University to capture occupant's variability in normal usage of their dorm room resources. Moreover, we include \textbf{seasonal dummy feature indicators} like time of day (morning vs. evening) and time of week (weekday vs. weekend) indicators. The intuition behind including such features is their ability to capture occupants' seasonal patterns in resource usage. Lastly, we incorporate pooled features that accurately model occupants variability in resources usage across each day. This results to \textbf{resources status continuous features}. Examples of such pooled features are: frequency of daily switches in resource status and percentage of resource usage across the day.

\section{Data Analysis}
The social game data set so obtained was filtered of possible inconsistencies in data. As our aim for this research was to segregate the behaviours of users based on their energy efficiency, we visualized the presence of latent clusters in the data. The data contains some independent features such as each resource's status, occupant's number of visits to the web portal etc. In addition, there are some dependent features such as gathered points and ranks. We only consider the independent features to apply the elbow method. 

\begin{figure}[!ht]
\includegraphics[height=5cm,width=0.6\textwidth]{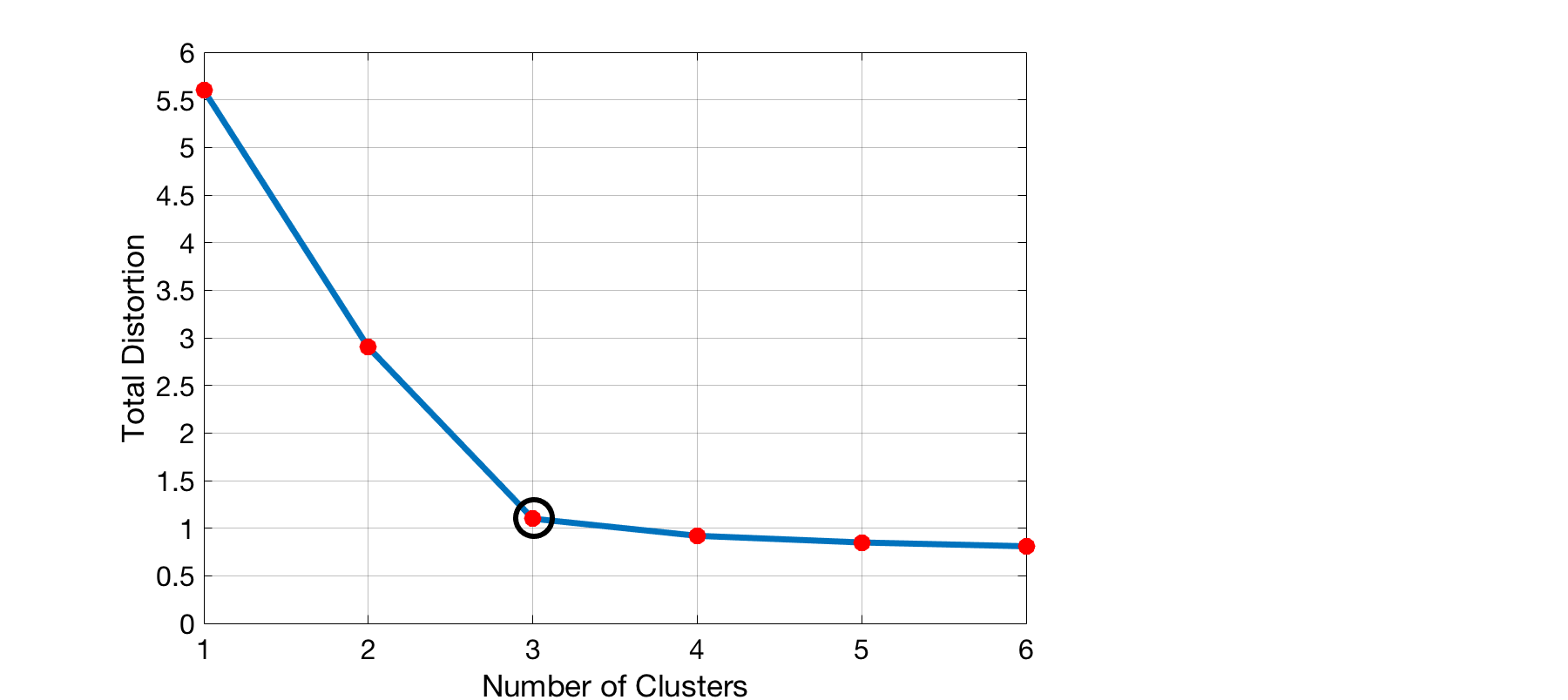}
\caption{Elbow Plot to choose optimal number of clusters}
\label{fig:elbow_plot}
\end{figure}

The elbow plot is given in Figure~\ref{fig:elbow_plot} and the elbow, marked by the circled point is achieved for optimal number of clusters as three. We calculate the Silhoutte score for each number of clusters, given in Table~\ref{tab:silhouette_scores}, which further strengthens the obtained result. Following the above analysis, we adopt both supervised and unsupervised methods to segregate the users energy usage behaviors.

\begin{table}[h!]
\centering
\begin{tabular}{ |c|c|c|c|c| } 
\rowcolor{Gray}
 \hline
 No. of Clusters & 2 & 3 & 4 & 5 \\ 
 \hline
 Silhouette Scores & 0.684 & 0.749 & 0.611 & 0.540 \\ 
 \hline
\end{tabular}
\caption{Silhouette Scores for different number of clusters}
\label{tab:silhouette_scores}
\end{table}

\subsection{Supervised segregation of players}\label{sec_supervised_segregation}
\begin{figure}[!ht]
    \includegraphics[height=5cm,width=0.48\textwidth]{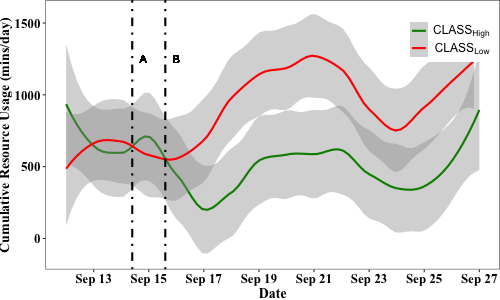}
    \caption{Variation of cumulative energy resource usage for players from high and low supervised classes}
    \label{fig:variational_energy_efficiency}
\end{figure}

Motivated by the fact that the optimal number of unsupervised clusters possible out of the social game data is three, we divided the users into three classes in a supervised way taking the ranks of the users as the benchmark. Let the target classes be represented by:
\begin{center}
[CLASS$_{low}$, CLASS$_{medium}$, CLASS$_{high}$]
\end{center}

where the subscripts signify the energy efficiency behavior of each class. Let the players be denoted by P$_1$, P$_2$, $\cdots$, P$_m$ and the data points corresponding to the ${i}^{th}$ player be d$^{i}_{1}$, d$^{i}_{2}$, $\cdots$, d$^{i}_{n_i}$. The whole range of ranks were divided into three equal segments, with the high, medium and low energy efficient rank groups being RANK$_{high}$, RANK$_{medium}$ and RANK$_{low}$ respectively. We assign the players to the classes as per the following formula, P$_i \in$ CLASS$_{X}$, where,
\begin{align}
    X = \underset{r \in [low,medium, high]}{argmax} \bigg\{\sum_{j=1}^{n_i}\mathds{1}[rank({d^{i}_{j}}) \in RANK_r]\bigg\}
\end{align}

where $\mathds{1}[\;\cdot\;]$ is the indicator function. So the output of above is allocation of each player in the whole spectrum of players into one of the three target classes. The behavior of a random player in a particular class, e.g. CLASS$_{high}$ represents the behavior of players showcasing high energy efficiency. Any algorithmic operation on each of the classes will provide results particular to players with corresponding energy efficiency. But, each player can exhibit periods of high as well as periods of medium and low energy efficiency maintaining the average performance as per standards of the class. Figure~\ref{fig:variational_energy_efficiency} shows the cumulative resource usage for players in CLASS$_{high}$ and CLASS$_{low}$, smoothened using LOESS with a 95\% confidence interval marked as the shaded region. Notice that during the period between lines marked A and B, a player from CLASS$_{high}$ has more resource usage than its counterpart in CLASS$_{low}$. This motivates to have data sample-centric segregation instead of player-centric segregation which requires unsupervised methods.

\subsection{Unsupervised segregation of data}\label{sec_unsupervised_segregation}
We use unsupervised clustering algorithms to segregate the data into clusters. We use Principal Component Analysis (PCA) to perform dimensionality reduction followed by Minibatch K-means algorithm with k= 3 to get three clusters out of the data. Let the clusters be represented by:
\begin{center}
[CLUSTER$_{1}$, CLUSTER$_{2}$, CLUSTER$_{3}$]
\end{center}
As the data is rich in energy efficiency behavior of players, the three clusters represent low, medium and high energy efficient samples. Then, we utilize the novelty of Graphical Lasso techniques to label the clusters as low/medium/high energy efficient with prior knowledge of behaviors exhibited by players in supervised classes.
\section{Graphical Lasso for Social Game}\label{graph_lasso}

We will now formulate a framework towards segmentation analysis that allows us to understand agent's decision-making model. Let the features representing the social game data set be denoted by the collection $Y = (Y_1, Y_2, \cdots, Y_S)$. $Y$ can be associated with the vertex set $V =  \{1, 2, \cdots, S\}$ of some underlying graph. The structure of the graph is utilized to derive inferences about the relationship between the features. We use the Graphical Lasso algorithm~\cite{mainbook} to realize the underlying graph structure. We assume the distribution of the random variables is Gaussian. The distribution for features is normalized since various features in the data have different scales and additionally the following analysis assumes them to have zero mean. As our aim is to learn the inherent relationship between the variables, normalization procedure on the data is well justified. 

Consider a random variable $Y_s$ at $s$ $\in$ $V$. We use the Neighbourhood-Based Likelihood for graphical representation of multivariate Gaussian random variables. Let the edge set of the graph be given by $\it{E}$ $\subset$ V$\times$V. The neighbourhood set of $Y_s$ is defined by 
\begin{align}
    \mathcal{N}(s) = \{k \in V | (k,s) \in \textit{E}\}
\end{align}
and the collection of all other random variables be represented by:
\begin{align}
    {Y}_{V\backslash\{s\}} = \{Y_{k}, k\in (V-\{s\})\}
\end{align}
For undirected graphical models, node for Y$_s$ is conditionally independent of nodes not directly connected to it given {Y}$_{\mathcal{N}(s)}$, i.e.
\begin{align}
    (Y_s|{Y}_{V\backslash\{s\}}) \sim (Y_s|{X}_{\mathcal{N}(s)})
\end{align}
The problem of constructing the inherent graph out of observations is nothing but finding the edge set for every node. This problem becomes predicting the value of $Y_s$ given Y$_{\mathcal{N}(s)}$, or equivalently, predicting the value of $Y_s$ given Y$_{\backslash\{s\}}$by the conditional independence property. The conditional distribution of $Y_s$ given Y$_{\backslash\{s\}}$ is also Gaussian, so the best predictor for $Y_s$ can be written as:
\begin{align}
    {Y}_{s} = {Y}_{V\backslash s}^T.{\beta}^{s} + {W}_{V\backslash s}
\end{align}
where W$_{V\backslash s}$ is zero-mean gaussian prediction error. The $\beta^{s}$ terms dictate the edge set for node s in the graph. We use $l_1$-regularized likelihood methods for getting a sparse $\beta^{s}$. Let the total number of data samples available be \textit{N}. The optimization problem is formulated as: corresponding to each vertex $s = 1, 2, \cdots ,S$, solve the following lasso problem:
\begin{align}
    \hat{{\beta}^{s}} \in \underset{\beta^s \in {\mathbb{R}}^{S-1}}{argmin} \bigg\{{\frac{1}{2N}\sum_{j=1}^{N}(y_{js}-{y^T_{j,{V\backslash s}}}}\beta^{s})^2 + \lambda{\|\beta^s\|}_{1}\bigg\}
\end{align} 
The implementation of Graphical Lasso algorithm is summarized in Appendix.\\

\section{Results}\label{results}

The results of the proposed segmentation analysis/learning method applied to data collected from the social game experiment is presented as follows.

\begin{figure*}
    { \centering                             
        \subfigure[Temporal Dependencies]{\includegraphics[width=0.33\textwidth]{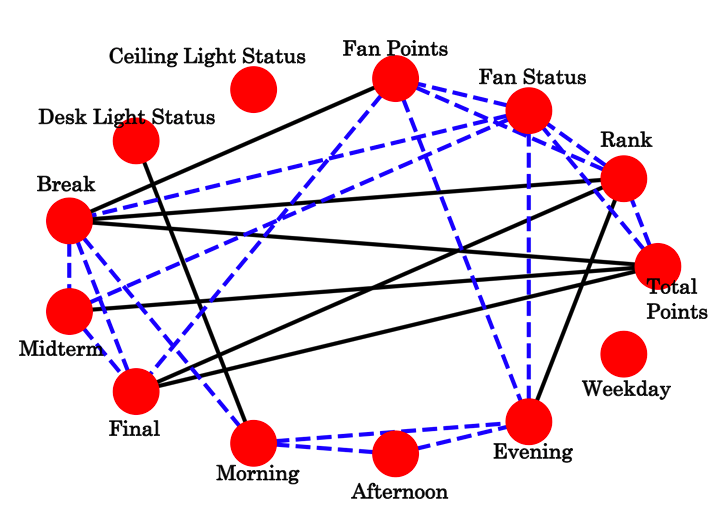}}
        \subfigure[External Factor Dependencies]{\includegraphics[width=0.33\textwidth]{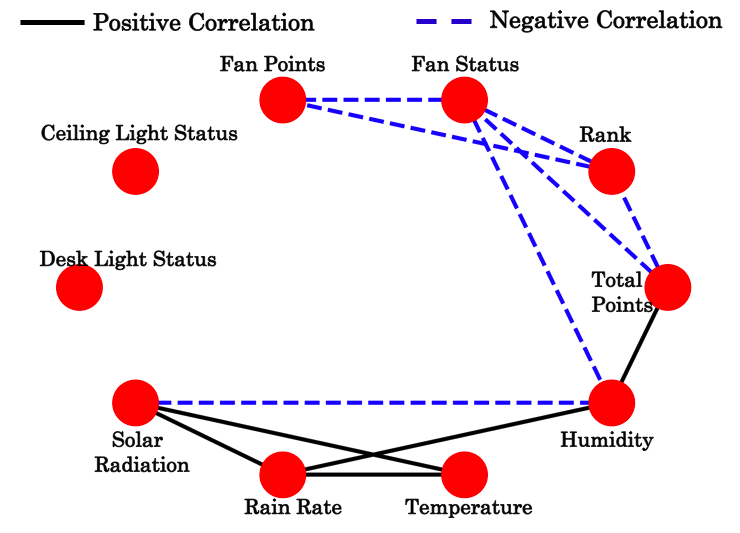}}
        \subfigure[Game Engagement Dependencies]{\includegraphics[width=0.33\textwidth]{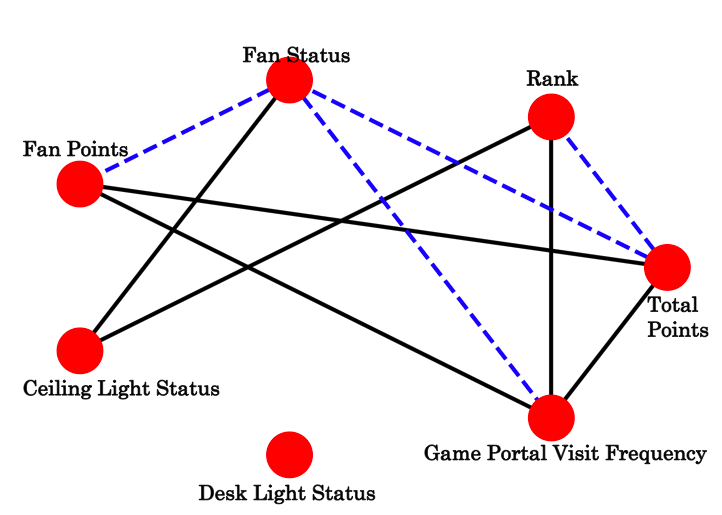}}
    }
    \caption{{\color{black}Feature correlations for a Low Energy Efficient Player ($\in$ CLASS$_{low}$)}}\label{fig_low}
    
    { \centering                             
        \subfigure[Temporal Dependencies]{\includegraphics[width=0.33\textwidth]{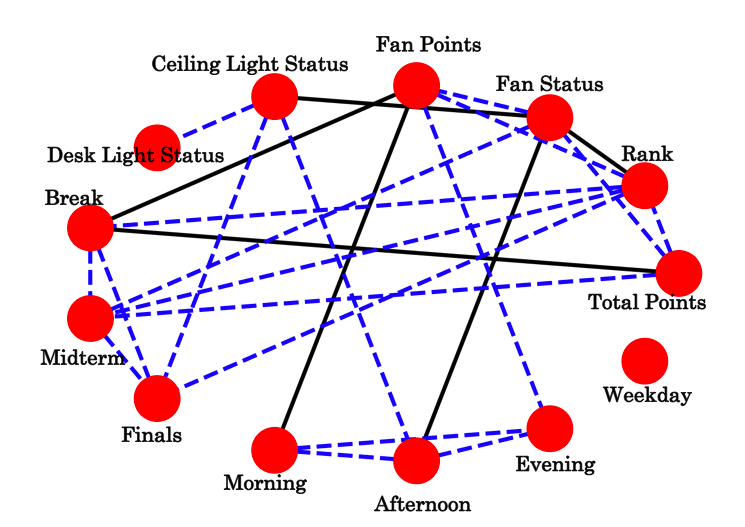}}
        \subfigure[External Factor Dependencies]{\includegraphics[width=0.33\textwidth]{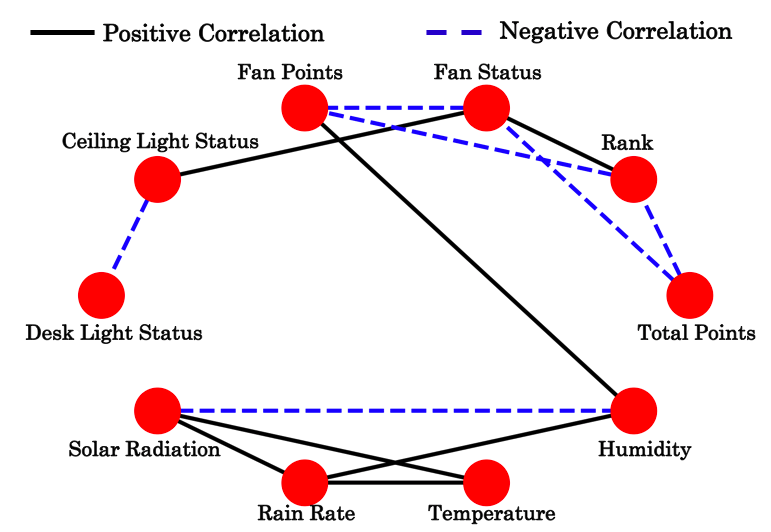}}
        \subfigure[Game Engagement Dependencies]{\includegraphics[width=0.33\textwidth]{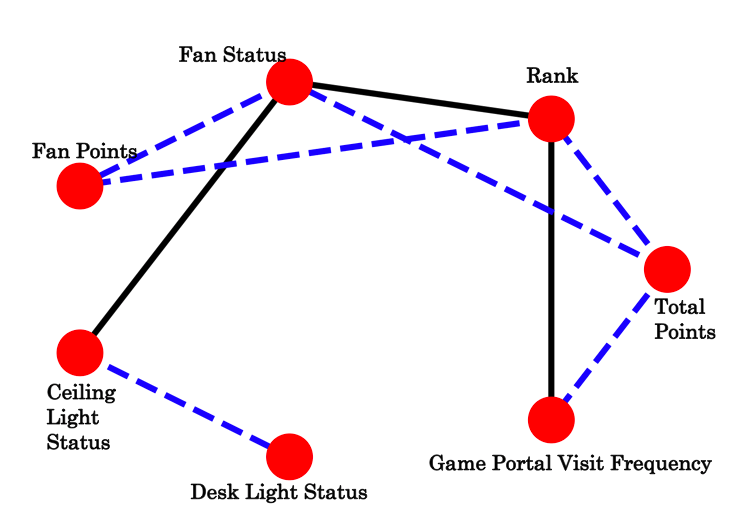}}
    }
    \caption{{\color{black}Feature correlations for a Medium Energy Efficient Player ($\in$ CLASS$_{medium}$)}}\label{fig_med}
    
    { \centering                             
        \subfigure[Temporal Dependencies]{\includegraphics[width=0.33\textwidth]{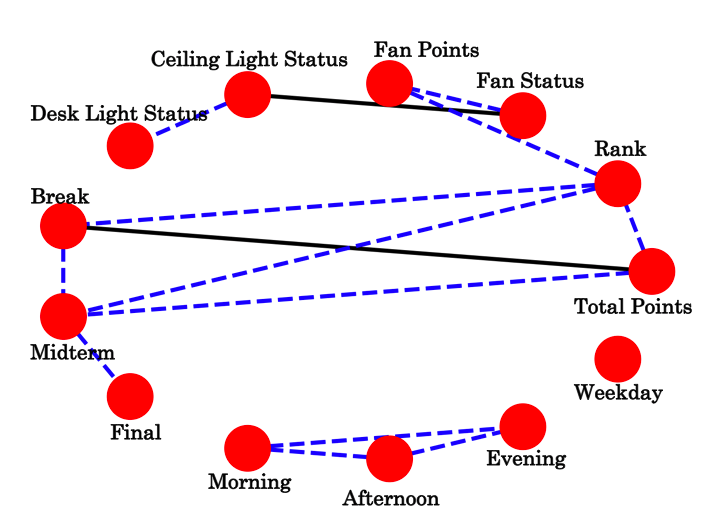}}
        \subfigure[External Factor Dependencies]{\includegraphics[width=0.33\textwidth]{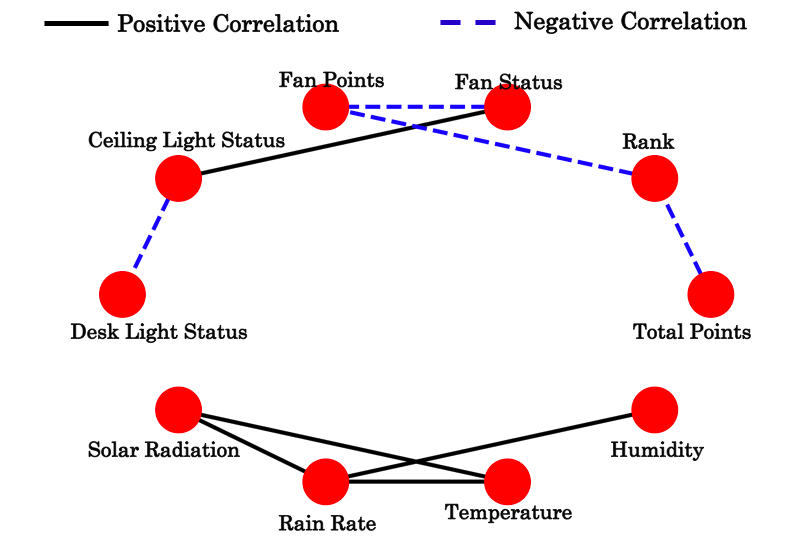}}
        \subfigure[Game Engagement Dependencies]{\includegraphics[width=0.33\textwidth]{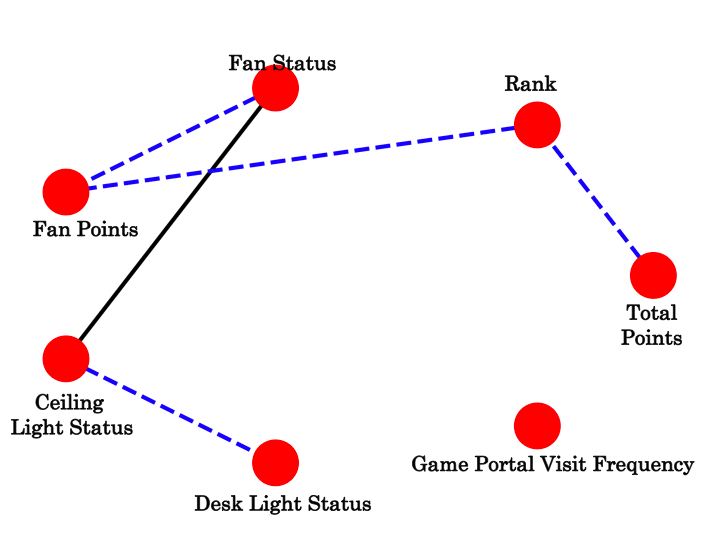}}
    }
    \caption{{\color{black}Feature correlations for a High Energy Efficient Player ($\in$ CLASS$_{high}$)}}\label{fig_high}
    
    { \centering                             
        \subfigure[Temporal Dependencies]{\includegraphics[width=0.33\textwidth]{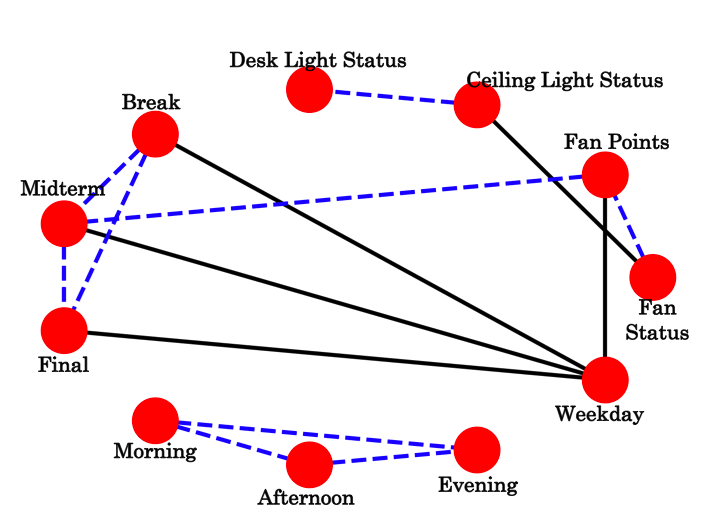}}
        \subfigure[External Factor Dependencies]{\includegraphics[width=0.33\textwidth]{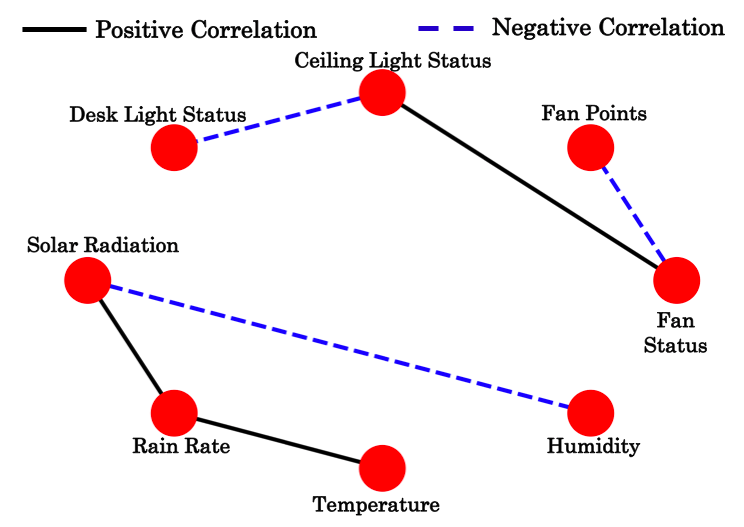}}
        \subfigure[Game Engagement Dependencies]{\includegraphics[width=0.33\textwidth]{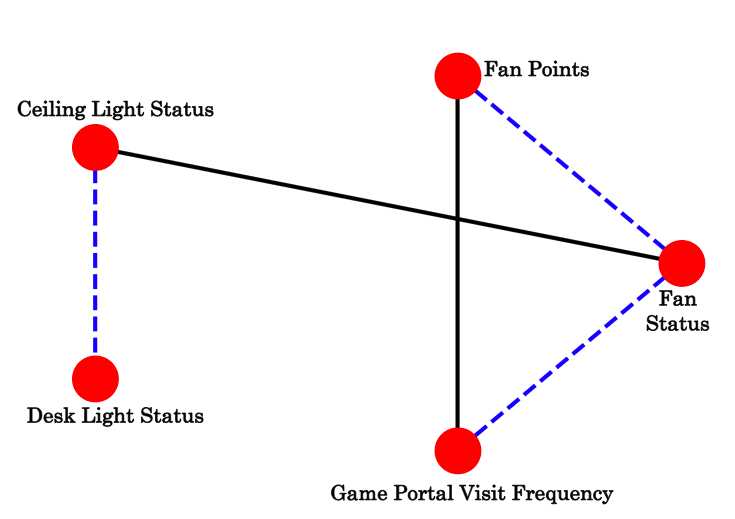}}
    }
    \caption{{\color{black}Feature correlations for energy usage behaviors in CLUSTER$_3$}}\label{fig_cluster3}
\end{figure*}

\subsection{Feature correlation learning in supervised segregation}\label{glasso_supervised}
We consider a representative player for each of the three classes obtained out of supervised segregation method described in Section~\ref{sec_supervised_segregation} to run graphical lasso and study the correlation between the features for that class. We group the features into different categories so as to study their influence on energy efficiency behaviors. Specifically, we consider \textit{Temporal} features like time of the day, academic schedules and weekday/weekends, \textit{External} features as outdoor temperature, humidity, rain rate etc. and \textit{Game Engagement} features like frequency of visits to game web portal.


The feature correlations for a low energy efficient player is given in Fig~\ref{fig_low}. A low energy efficient player in the setting of the game tries to use each resource independently which can be observed in Figure~\ref{fig_low}(a) with no correlation between the corresponding resource usage identifiers. There is a positive correlation between morning time and desk light usage indicating somewhat heedless behavior towards energy savings. The absolute energy savings increase during the breaks and finals, given by positive correlation with total points, but it is not significant as compared to the amount of energy savings that other players in the game exhibit during the same period, thus increasing the rank. External parameters play a significant role in energy usage behavior of this class. The operation of the ceiling fan is driven by external humidity as given in Figure~\ref{fig_low}(b). Figure~\ref{fig_low}(c) indicates that their frequency of visits to the game web portal is motivated by sub-optimal performance in the game.

Feature correlations for a medium energy efficient player is given in Fig~\ref{fig_med}. The player showcases predictable behaviors with correlations between desk light, ceiling light and ceiling fan usage (Figure~\ref{fig_med}(a)). The player co-optimizes the usage by alternating the use of ceiling and desk light. Different occasions like break, midterm and final are marked by energy saving patterns. Unlike a low energy efficient player, the player in this class tries to save energy in a conscious manner shown by reduced fan usage during the morning and reduced light usage during the afternoon. The fan usage is influenced by the external humidity, shown by Fig~\ref{fig_med}(b). The game engagement patterns for a player in this class (Fig~\ref{fig_med}(c)) is similar to that of the low energy efficient class.

Fig~\ref{fig_high} shows the feature correlations for a high energy efficient player. This player also exhibits predictable behaviors as showcased by a player in medium energy efficient class. Opportunistically, this player saves energy during breaks and midterms as shown by negative correlation between the corresponding flags and rank in Figure~\ref{fig_high}(a). Notice that there exists a negative correlation between midterm flag and total points, indicating decrease in absolute amount of points. However, the points are still higher than the points by other players which marks improvement in the rank. This behavior is completely opposite to what is  exhibited by a player in low energy efficient class. The player is neither affected by the time of the day, nor by the external factors (Figure~\ref{fig_high}(b)) showing a dedicated effort to save energy. The game engagement behavior for this player, given in Figure~\ref{fig_high}(c) is inconclusive, possibly because it is dominated by other energy saving factors.
\subsection{Labelling unsupervised clusters using feature correlations in supervised segregation}
We applied graphical lasso to CLUSTER$_1$,CLUSTER$_2$ and CLUSTER$_3$ obtained from unsupervised segergation of data in Section~\ref{sec_unsupervised_segregation}. Based on the feature correlation knowledge gained from different classes in Section~\ref{glasso_supervised}, we labelled the clusters as having low, medium or high energy efficient data. The feature correlations for CLUSTER$_3$ is shown in Fig~\ref{fig_cluster3}. It is evident from Fig~\ref{fig_cluster3}(a) that data in CLUSTER$_3$ exhibit predictability in behavior with correlations between resource usage flags. Also the weekdays are marked by energy savings in terms of increase in fan points. The time of the day is also unrelated to the performance. Neither do the external factors contribute to the performance (Figure~\ref{fig_cluster3}(b)). The engagement in the game also boosts the points (Figure~\ref{fig_cluster3}(c)).

All the above behaviors are indicative of the similarity between the energy efficiency characteristics manifested by CLUSTER$_3$ and the high energy efficient class obtained using supervised segregation. So, CLUSTER$_3$ is labelled as the high energy efficient cluster. Following the same comparison, CLUSTER$_1$ and CLUSTER$_2$ are labelled as the medium and low energy efficient clusters respectively. 

\begin{figure}
    { \centering                             
        \subfigure[]{\includegraphics[height=5cm,width=0.3\textwidth]{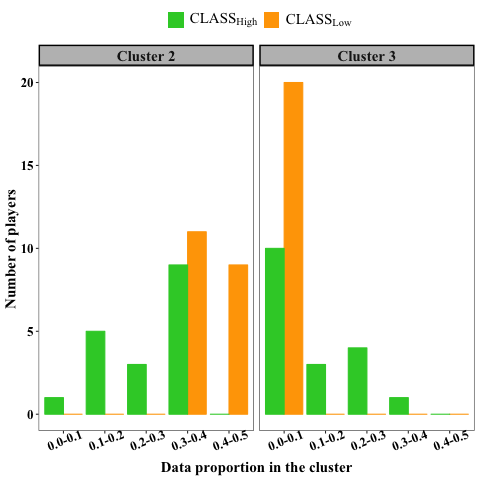}}
        \subfigure[]{\includegraphics[height=5cm,width=0.16\textwidth]{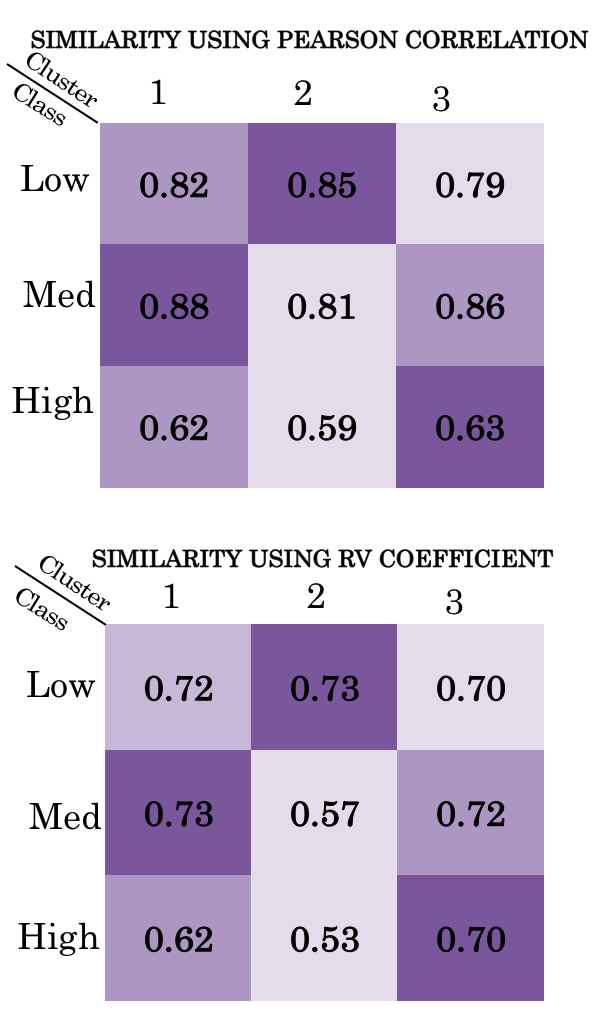}}
    }
    \caption{{\color{black}(a) Distribution of high and low energy efficient players data proportion in various clusters\\(b) Similarity between feature correlation matrices}}\label{fig_bar_corr_matrix}
\end{figure}

To further strengthen our inference, we compute the proportion of data each player from the three supervised classes has in each of the three unsupervised clusters. The bar chart showing number of players in each data proportion bucket for a cluster is shown in Fig~\ref{fig_bar_corr_matrix}(a). It is evident from the bar chart that majority of players from low energy efficient class have substantial amount of data in CLUSTER$_2$ and minimal amount of data in CLUSTER$_3$. On the other hand, though the number of players for each data proportion bucket for players from high energy efficient class is somewhat scattered, it is more inclined towards higher value in CLUSTER$_3$. Finally, Fig~\ref{fig_bar_corr_matrix}(b) shows the similarity between the feature correlation matrices computed using Pearson Correlation and RV coefficient for matrices. It represents the match as: \{CLUSTER$_1$ $\sim$ CLASS$_{medium}$\},\{CLUSTER$_2$ $\sim$ CLASS$_{low}$\} and \{CLUSTER$_3$ $\sim$ CLASS$_{high}$\}. All these corroborate our earlier proposition about labelling the clusters as low, medium or high energy efficient.
\subsection{Causal Relationship between features}

\begin{table*}[!ht]
  \centering
  \resizebox{2.1\columnwidth}{!}{
  \begin{tabular}{|c|c|c|c|c|c|c|c|c|c|c|c|c|c|c|c|}
  \rowcolor{Gray}
  \hline
    \multicolumn{1}{|c|}{Test whether $X$ causes $Y$} & \multicolumn{2}{c|}{Fan $\implies$ Ceiling Light} & \multicolumn{2}{c|}{Humidity $\implies$ Fan} & \multicolumn{2}{c|}{Desk Light $\implies$ Fan} & \multicolumn{2}{c|}{Ceiling Light $\implies$ Desk Light} & \multicolumn{2}{c|}{Morning $\implies$ Desk Light} & \multicolumn{2}{c|}{Afternoon $\implies$ Fan} & \multicolumn{2}{c|}{Evening $\implies$ Ceiling Light} \\ \hline
    \rowcolor{Gray}
Player type & p-value & F-statistic & p-value & F-statistic & p-value & F-statistic & p-value & F-statistic & p-value & F-statistic & p-value & F-statistic & p-value & F-statistic\\ \hline
Low Energy Efficient & 0.54 & 0.37 & \cellcolor{blue!25}\textbf{0.004} & 8.12 & 0.06 & 3.55 & 0.81 & 0.06 & 0.4 & 0.71 & \cellcolor{blue!25}\textbf{0.01} & 6.1 & \cellcolor{blue!25}\textbf{0} & 25.3\\ \hline
Medium Energy Efficient & \cellcolor{blue!25}\textbf{0} & 21.2 & \cellcolor{blue!25}\textbf{0.008} & 7.06 & \cellcolor{blue!25}\textbf{0} & 113.6 & \cellcolor{blue!25}\textbf{0} & 25.8 & 0.23 & 1.41 & 0.46 & 0.55 & \cellcolor{blue!25}\textbf{0.0007} & 11.5\\ \hline
High Energy Efficient & \cellcolor{blue!25}\textbf{0} & 21.9 & 0.12 & 2.36 & 0.99 & 0.003 & 0.93 & 0.007 & 0.63 & 0.22 & \cellcolor{blue!25}\textbf{0.04} & 4.2 & 0.52 & 0.41\\ \hline
\end{tabular}}
\caption{Causality test results among various potential causal relationships. In bold are the p-values (shaded in blue) in cases that Granger causality is established through F-statistic test between features. p-values lower than $0.05$ indicate strong causal relationship in 5\% significance level}
\label{tab:grangers_causality}
\end{table*}
 The results for causal relationship between features using Granger causality test is given in Table~\ref{tab:grangers_causality}. Under null hypothesis $H_0$, $X$ does not Granger-cause $Y$. So, a p-value lower than $0.05$ (5\% significance level) indicates a strong causal relationship between the tested features and implies rejecting the null hypothesis $H_0$.

The p-values (shaded in blue) for which Granger causality is established are highlighted in the table. Interestingly, for medium and high energy efficient building occupants, ceiling fan usage causes ceiling light usage. This fact in turn confirms the predictive behavior for them as mentioned earlier. In both low and medium energy efficient building occupants, external humidity causes ceiling fan usage. This is an indicator that their energy usage is affected by external weather conditions. However, for high energy efficient building occupants external humidity doesn't cause ceiling fan usage. This shows that they are highly engaged with the proposed gamification interface and try to minimize their energy usage. Another interesting result is that the evening label causes ceiling light usage for both low and medium energy efficient building occupants. But this is not the case for high energy efficient building occupants, for whom ceiling light usage is better optimized as a result of their strong engagement with the ongoing social game, eventually leading to exhibition of better energy efficiency.

\subsection{Energy savings through gamification}

\begin{figure}[!ht]
\centering
  \includegraphics[width=0.45\textwidth]{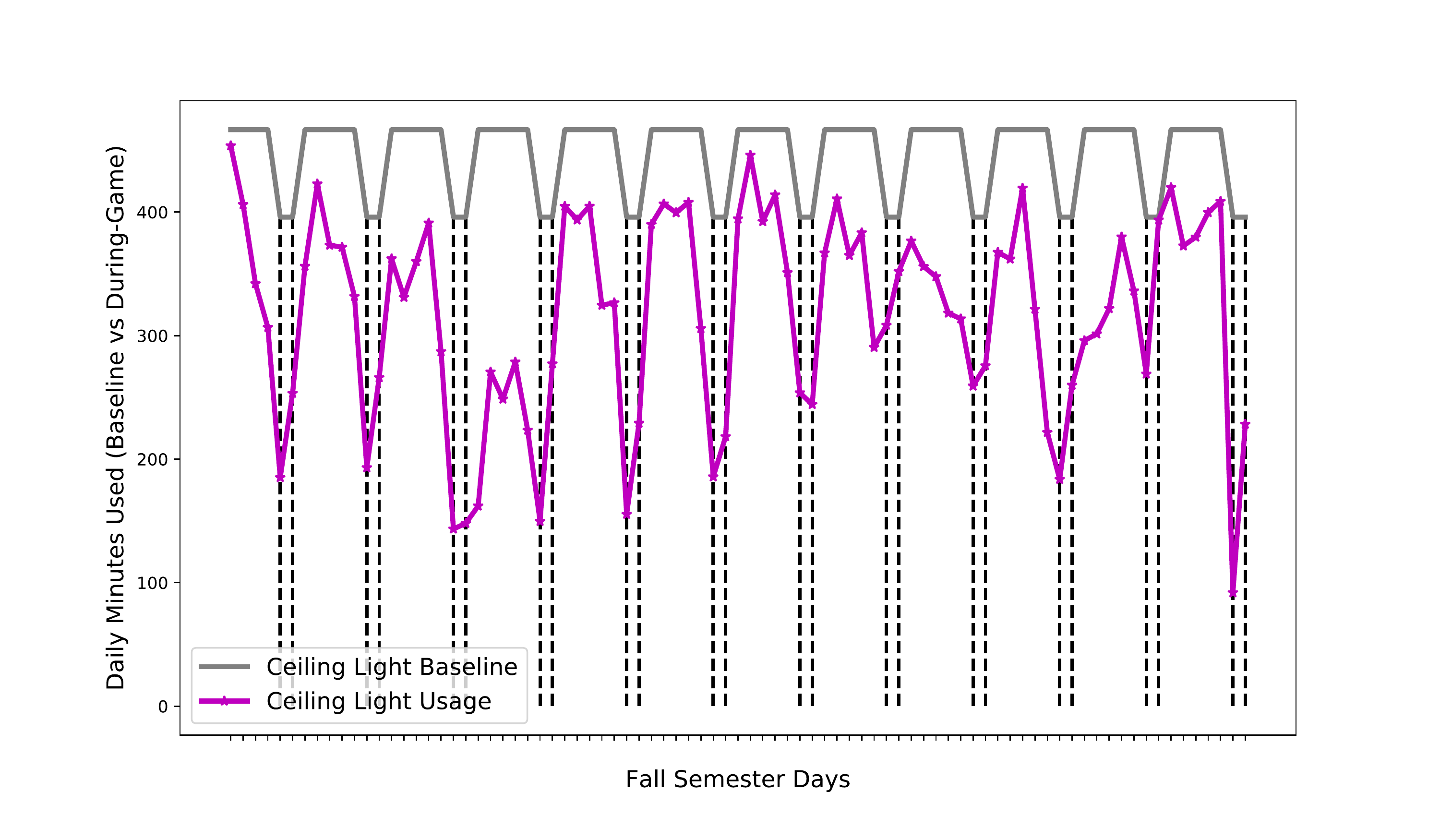}
  \caption{Fall semester daily average minutes ceiling light usage compared to weekday \& weekend average baselines. Vertical black dashed lines indicate a weekend period}
  \label{fig:Fall_Energy}
\end{figure}

Lastly, we present the achieved energy savings in Fall semester version of the social game. In Figure~\ref{fig:Fall_Energy} we present the daily average minutes ceiling light usage compared to weekday \& weekend average baselines.\footnote{Weekday \& weekend average baselines are computed using past usage data over a period of four weeks before the beginning of the social game.}The region between two vertical black dashed lines indicate a weekend period, which has a different average baseline target for the occupants. In Table~\ref{tab:fall_hyp}, we see the hypothesis testing values for different devices.  In the table, the \textquotedblleft{Before}\textquotedblright column denotes the data points gathered before the game was officially started. The \textquotedblleft{After}\textquotedblright column is the data during the game period. Data points in the tables are shown for both weekday and weekend periods and represent the average usage of all the occupants. Usage is defined in minutes per day. In all cases of the devices, we have a significant drop in usage between the two periods.\footnote{Drop in usage (column named $\Delta$ \%) indicates reduction in the average usage of all occupants.}Resulting p-values from 2-sample t-tests show highly significant change in usage patterns. Moreover, we can see a much larger drop in usage is achieved over the weekends. These are significant results showing how we can optimally incentivize occupants in residential buildings to reduce the energy usage.


\begin{table}[h!]
\centering
\resizebox{\columnwidth}{!}{%
\begin{tabular}[!h]{ |p{0.95cm}|p{0.8cm}|p{0.775cm}|p{0.775cm}|p{0.65cm}|p{0.8cm}|p{0.775cm}|p{0.775cm}|p{0.65cm}|}

\hline 
\rowcolor{Gray}   & \multicolumn{4}{c|}{Weekday} & \multicolumn{4}{c|}{Weekend}\\
 \hline
 \rowcolor{Gray}
 Device & Before & After & $p$-value & $\Delta$ \% & Before & After & $p$-value & $\Delta$ \%\\
 \hline
Ceiling Light & 417.5 & 393.9 & 0.02 & 5.6 & 412.3 & 257.5 & 0 & 37.6\\
 \hline
 Desk Light & 402.2 & 157.5 & 0 & 60.8  & 517.6 & 123.3 & 0 & 76.2\\
 \hline
Ceiling Fan & 663.5 & 537.6 & 0 & 19.0  & 847.1 & 407.0 & 0 & 51.9\\ 
\hline
\end{tabular}
}
\caption{Fall Game (Before vs After) minutes per day usage hypothesis testing}
\label{tab:fall_hyp}
\end{table}

\newpage
\section{Conclusions}
We conducted a smart building social game where occupants concurrently optimized their room's energy resource usages and participated in a lottery. A general framework for segmentation analysis in human-centric cyber-physical systems was presented. The analysis included clustering of agents behaviors and an explainable statistical model representing the contributed features motivating their decision-making. To strengthen our results, we examined several feature correlations using granger causality test for potential causal relationships. 

Our ultimate goal for the segmentation analysis is to develop a gamification methodology, which simultaneously learns occupant preferences while also opening avenues for feedback. The latter is important for integrating explainable statistical models with adaptive control or online incentive design. Not surprisingly, static programs for encouraging energy efficiency are less efficient with passing of time~\cite{laitner:2000aa,schipper:2000aa}. Hence, by leveraging proposed segmentation analysis we can create an adaptive model that learns how users' preferences change over time, and thus generate the appropriate incentives to ensure active participation. Furthermore, the learned preferences can be adjusted through incentive mechanisms~\cite{ratliff2018adaptive} to enact improved energy efficiency.
\section{Acknowledgments}
This research is funded by the Republic of Singapore's National Research Foundation through a grant to the Berkeley Education Alliance for Research in Singapore (BEARS) for the Singapore-Berkeley Building Efficiency and Sustainability in the Tropics (SinBerBEST) Program. BEARS has been established by the University of California, Berkeley as a center for intellectual excellence in research and education in Singapore. The work of I. C. Konstantakopoulols was supported by a scholarship of the Alexander S. Onassis Public Benefit Foundation.

\clearpage
\bibliographystyle{unsrt}
\small{
\bibliography{glasso}}

\clearpage
\appendix
\begin{tikzpicture}[yscale=3] 
\draw [line width=0.65mm, black ] (0,-1) -- (8.5,-1) node [right]{};;
\end{tikzpicture}
 Algorithm 1: GRAPHICAL LASSO ALGORITHM FOR \\ GAUSSIAN GRAPHICAL MODELS\\
\begin{tikzpicture}[yscale=3] 
\draw [line width=0.65mm, black ] (0,0) -- (8.5,0) node [right]{};;
\end{tikzpicture}
\begin{enumerate}[1]
\item For vertices $s = 1, 2, \cdots ,S$:
\begin{enumerate}[(a)]
    \item Calculate initial loss ${\|Y_{s} - Y^T_{V\backslash s}\beta^{s}\|}^{2}_{2}$
    \item Untill Convergence:
    \begin{enumerate}[i]
        \item Calculate partial residual $r^{(s)}$ = $Y_s$ - $Y^T_{V\backslash s}\beta^{s}$
        \item For all j $\in {V\backslash s}$, Get ${\beta}^{s,new}_j$ = ${\textbf S_{\lambda}\big(\frac{1}{N}\langle r^{(s)},Y_j\rangle\big)}$ 
        \item Compute new loss = ${\|Y_{s} - Y^T_{V\backslash s}\beta^{s,new}\|}^{2}_{2}$
        \item Update ${\beta^{s}}$ = ${\beta}^{s,new}$
    \end{enumerate}
    \item Get the neighbourhood set $\mathcal{N}$(s)= supp(${{\beta}}^{s}$) for s
\end{enumerate}
\item Combine the neighbourhood estimates to form a graph estimate ${G}$ = (V, $\textit{E}$) of the random variables.
\end{enumerate}
\begin{tikzpicture}[yscale=3] 
\draw [line width=0.65mm, black ] (0,-1) -- (8.5,-1) node [right]{};;
\end{tikzpicture}

$\textbf S_{\lambda}(\theta)$ is soft thresholding operator as $sign(\theta)(|\theta|-\lambda)_{+}$.\\
\newpage
\hspace{50mm}

The penalty factor $\lambda$ determines the sparsity in the regression coefficients vector for a node. Ideally, the coefficient matrix must be accurate enough for regression while maintaining notable amount of sparsity. For optimal design of $\lambda$ in Graphical Lasso run for a vertex s, we take 10 values in logarithmic scale between $\lambda_{max}$ and $\lambda_{min}$ as given below and conduct a line search to find the penalty factor which brings the minimum loss.
\begin{align}
 \lambda_{max} = \frac{1}{N}\underset{j\in V\backslash s}{max}|\langle Y_j,Y_s\rangle|
 \end{align}
 \begin{align}
 \lambda_{min} = \frac{\lambda_{max}}{100}
\end{align}
Implementing a coordinate descent approach~\cite{mainbook} the time complexity of the proposed algorithm in $O(pN)$ for a complete run through all $p$ features. We also do 5-fold cross validation to ensure accurate value of the coefficients $\beta^{s}$. Use of partial residuals for each node significantly reduces the time complexity of the algorithm.
\end{document}